\documentclass[10pt, conference]{IEEEtran}
\usepackage{graphicx}
\usepackage{cite}
\usepackage{amsmath}
\usepackage{epstopdf}
\usepackage{subfig}
\usepackage[section]{placeins}
\usepackage{xcolor}
\usepackage{algcompatible}
\usepackage{algorithm}

\providecommand{\keywords}[1]{\textbf{\textit{Keywords:}} #1}

\IEEEoverridecommandlockouts
 \usepackage[pscoord]{eso-pic}

\begin{document}
\IEEEoverridecommandlockouts
\IEEEpubid{\makebox[\columnwidth]{978-1-6654-4316-6/21/\$31.00 \copyright 2021 IEEE \hfill} \hspace{\columnsep}\makebox[\columnwidth]{ }}
\title{Toward Digitalization: A Secure Approach to Find a Missing Person Using Facial Recognition Technology}
 \author{
\IEEEauthorblockN{Abid Faisal Ayon}
\IEEEauthorblockA{\textit{Department of Computer Science and Engineering} \\
\textit{Bangladesh University of Business and Technology}\\
Dhaka, Bangladesh \\
16172103440@cse.bubt.edu.bd}

\and 

\IEEEauthorblockN{S M Maksudul Alam}
\IEEEauthorblockA{\textit{Department of Computer Science and Engineering} \\
\textit{Bangladesh University of Business and Technology}\\
Dhaka, Bangladesh \\
maksudulmimon@bubt.edu.bd}
}
\maketitle

\begin{abstract}
Facial Recognition is a technique, based on machine learning technology that can recognize a human being analyzing his facial profile, and is applied in solving various types of real-world problems nowadays. In this paper, a common real-world problem -- finding a missing person -- has been solved in a secure and effective way with the help of facial recognition technology. Although there exist a few works on solving the problem, the proposed work is unique with respect to its security, design, and feasibility. Impeding intruders in participating in the processes and giving importance to both finders and family members of a missing person are two of the major features of this work. The proofs of the works of our system in finding a missing person have been described in the result section of the paper. The advantages that our system provides over the other existing systems can be realized from the comparisons, described in the result summary section of the paper. The work is capable of providing a worthy solution to find a missing person on the digital platform.

\end{abstract}

\keywords{Secure approach; Administrative process; Cross-Matching; Facial recognition; Facial detection; Facial Mapping; Deep learning}

\section{Introduction}\label{sec:intro}
\noindent

 Facial recognition\cite{r1} is an approach that uses different techniques such as \textit{PCA}, \textit{LBPH}, \textit{Deep Learning}\cite{r3}, \textit{SVM}, \textit{MVFD}, etc. to find out the similarity of human faces depending on different facial characteristics. As human beings are different with respect to the characteristics of their facial constituents and the overall structure, identifying a person analyzing his personal facial characteristics can be performed effectively.\\
 In the past few years, many interesting works have already been performed in the field of facial recognition. For instance, Derya Ozkan and Pınar Duygulu et al\cite{n2} presented a method for finding people out by investigating the photographs and videos using a face recognition system. Debayan Deb, Divyansh Aggarwal,  Divyansh and Jain, Anil K et al.\cite{n3} proposed a system to find missing children using \textit{aging deep face features}.
 Natsume, Ryota, and Yatagawa, Tatsuya and Morishima, Shigeo et al.\cite{L12} proposed a method of using latent variables of facial landmarks instead of facial textures to perform image-based face-swapping. Moreover, Raghuwanshi, Anshun and Swami, Preeti D et al.\cite{L13} developed a system to maintain classroom attendance automatically using video-based facial recognition. Nech, Aaron, and Kemelmacher-Shlizerman, Ira et al.\cite{L11}, in their work, trained, tested, and compared the results of different facial recognition models with respect to the same large-scale dataset. \\
 People of various ages are getting lost because of their immaturity, abnormality, or unintentional accidents. Depending on different conditions, a missing person may not be able to express his destination rightly. The family members of a missing person have to go through immense hardship in finding the missing person. At the same time, a finder, after finding a missing person, has to face a great hassle while helping a missing person to reach his family. Moreover, Sometimes people with unscrupulous mentalities can try to take some advantage using the information of a missing person. Thus, the hardships of families of a missing person get intensified. Generally, people are habituated to solve the problem manually following some conventional, naive approaches such as recording missing diaries to the police station, advertising missing news on public media, and so on. However, the rules and regulations that are involved in the conventional approaches, in most cases, pose unbearable hassles to the stakeholders of the problem. Other drawbacks of the conventional approaches are their vulnerability and time consumption while providing a standard solution to the problem. This is why solving the problem -- finding a missing person -- technologically ensuring security and feasibility is not only challenging but also very demanding in the modern age.\\
In our work, we have proposed a secure, nation-wise connected, centralized system basing on facial recognition technology that can be helpful in dealing efficiently and securely with the problem. The system serves a finder and a family member of a missing person with equal importance and suitable services breaking the restrictions such as places and times. The work focuses on the security aspects regarding the system with high gravity. Security issues that can be adversarial to either the stakeholders or the system have been investigated and handled with appropriate concerns. The processes, performed inside the system are impermeable to the intruders. Information along with the photos uploaded in our system are verified thoroughly in multiple steps to overcome the security issues. Again, the system maintains well-organized databases that are helpful in preserving efficiency while dealing with data on large scale. The system can notify the concerns within the shortest possible time on the success of finding a missing person. 
The major \textit{contributions} of the work can be listed as follows:
\begin{itemize}
\item Security to the processes of finding a missing person.
\item Flexibility to both sides of the stakeholders: family members and finder of a missing person.
\item Notification to the concerns instantly while a missing person is identified
\end{itemize}

The remainder of the paper is organized in the segments as follows. In Section~\ref{sec:prev}, some of the major related research works have been described briefly. Our proposed system has been described in detail in Section~\ref{sec:algo}. In Section~\ref{sec:result}, the results, and in Section~\ref{sec:resultsummary}, the result-summary of our work have been shown. Our future work has been described in Section~\ref{sec:fw}. The concluding remarks can be found in Section~\ref{sec:conclusion}.  

\section{Previous works}\label{sec:prev}
There have been a few research works bases on finding a missing person using facial recognition technology. Some of the major works and their pitfalls can be investigated as follows:\\
Xin Jin, Shiming Ge, Chenggen Song, Xiaodong Li, Jicheng Lei, Chuanqiang Wu, and Haoyang Yu et al.\cite{L1} proposed a system, ``Double-blinded finder: A two-side privacy-preserving approach for finding missing children". In their work, the privacy of a missing child is ensured by using their cipher images instead of their original ones in public clouds and maintaining public-private keys. The public key-pair (pkF and pkI) is generated from the feature vector of a missing child image. However, in the work, the information of the kin of a missing person has not been verified and there exists a backdoor for an intruder to pretend like kin of a missing person.\\
Bharath Darshan Balar, D S Kavya, Chandana M, Anush E, Vishwanath R Hulipalled et al.\cite{Ex} presented ``Efficient Face Recognition System for Identifying Lost People". The system in their work saves a missing person's photo along with information uploaded either by police or the family of the person in the database and notifies them when the photo gets matched with a photo uploaded by a finder later. However, their system performs a missing person's kin-side matching partially and does not care about any verification or security issue. Their system cannot handle \textit{Case 3} and \textit{Case 4}, mentioned in the result section. \textit{Case 2}  has been handled partially in their system. Along with sharing a common database for both of the found and missing people images, their work is unable to deal with a crucial aspect -- multiple entries for the same missing person.\\
Peace Muyambo et al.\cite{n1} worked on ``An Investigation on the Use of LBPH Algorithm for Face Recognition to Find Missing People in Zimbabwe". In their work, they used the Local  Binary Patterns  Histograms(LBPH) Algorithm to perform facial recognition for finding missing people in Zimbabwe. The accuracy of their work was proportional to the number of images, their system was trained with. The work is not secure enough to be used heedlessly.
\section{Proposed System}\label{sec:algo}
\noindent
In this paper, a secure, efficient, and pragmatic system has been proposed to solve a problem -- finding missing persons -- with the help of modern technology: facial recognition. Providing security to find a missing person using an online platform is the very first motivation of our work. The system maintains considerable security, efficiency, and flexibility while providing services.
\textit{Our proposed system} can defeat the traditional and existing systems of finding a missing person with its efficiency, completeness, security, and feasibility. The advantages of our proposed system can be listed as follows:
\begin{itemize}
\item People do not need to go through any complex official rules and regulations 
\item There is no chance to be duped by intruders while using the system
\item Security of a missing person is ensured
\item The steps that need to be followed are more flexible and convenient to the users.
\item The efficiency of the system is sustained while dealing with data on a large scale
\end{itemize}
In the rest of the section, our proposed system will be described in detail with the descriptions of each of the sub-systems, their workflows, and the algorithms they follow.

\subsection{\textbf{Interaction diagram of the proposed System:}}
\begin{figure}[tbh!]
    \frame{\includegraphics[width=3.5in]{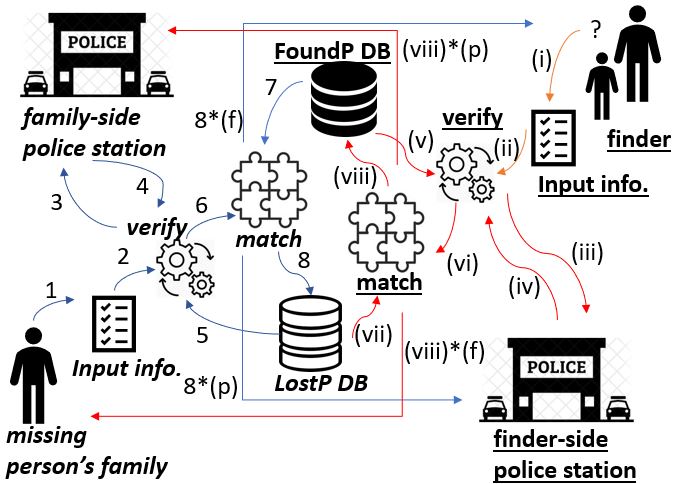}}
\caption{Interaction diagram of the system}
\label{fig:iD}
\end{figure}

The interactions of a finder and a family member of a missing person with our system can be depicted by the interaction diagram in Fig.~\ref{fig:iD}. 

From the interaction diagram, it can be seen that the two subsystems -- the missing person's kin-side subsystem and the finder-side subsystem -- are maintaining two separate databases:``FoundPeople" and ``LostPeople", respectively, and connected by the cross-matching. The user-uploaded information and photos are verified thoroughly before proceeding toward further approaches in the system. At first, the information, except the photo uploaded by a user, is verified by an appropriate and authorized police station keeping the user's status as \textit{AP} (Administrative Processing) so that an intruder cannot get into the process. After getting approval from the police station, the status of the user is updated and the uploaded photo is saved after performing another verification step. Verification of the uploaded photo is also important for the system's security. It is not expected to create multiple entries for a missing person at a time. Intruders can use a large number of same and valid entries to commence system failure. While matching the uploaded image from the missing person's kin and the finder, the system uses the images from ``FoundPeople" and ``LostPeople" databases respectively. Here the cross-matching occurs. Depending on the result of matching, either the information along with the photos are saved in the suitable database, or the system shows and notifies the concerns.

\subsection{\textbf{Proposed Algorithm}}
The well-defined tasks, performed in the proposed system are secure enough, easy to follow and effective in producing the desired outcomes. The steps of the algorithms can be summarized as follows:
\begin{description}
\item [\textbf{Step 1:}] \hspace{.2cm} Get  input  a  recognizable  photo  of  a  missing  person along  with  the  required  information  from  the  uploader
\item [\textbf{Step 2:}] \hspace{.2cm} Verify the uploaded photo and the information in the  system
\item [\textbf{Step 3:}] \hspace{.2cm} Save the verified encoded image and  uploaded information in the database
\item [\textbf{Step 4:}] \hspace{.2cm} Match  the  recently  encoded  photo with the encoded unrecognized photos that are saved in cross-directory of the system
\item [\textbf{Step 5:}] \hspace{.2cm} Show the matching result to the uploader
\item [\textbf{Step 6:}] \hspace{.2cm} Notify the appropriate person
\item [\textbf{Step 7:}] \hspace{.2cm} Notify the police  station
\end{description}

The steps of the algorithm are detailed in the following discussion:

\subsubsection{\textbf{Get input a recognizable photo of a missing person along with the required information from the uploader}}
A recognizable photo of a missing person and valid information i.e., national identity (\textit{NID}) of the user, contact number, email address, police station information, location, etc. are uploaded to the system so that the system can perform its \textit{match} and \textit{notify} functionalities. The status of the user's process is kept \textit{AP} until the verification step is fully complete in the system.

\subsubsection{\textbf{Verify the uploaded photo and the information in the system}}
At first, the authenticity of the user-given police station's email address is verified. If the police station email address is valid, then the information is sent to the police station email for verification and recording a missing diary. As \textit{NID}, phone number, etc. are registered officially for all the citizens of a country, police can easily verify the user's information and the missing case. If police approve the missing case and the information of the user, the uploaded photo of the missing person is verified. The photo verification is performed so that the entries in the system do not get repeated. If any anomaly is detected while verification, the system does not let the user proceed in the further steps.

\subsubsection{\textbf{Save the verified encoded image and uploaded information in the database}}
The encoded version of recently uploaded photo of the missing person and the information of the user is stored in appropriate system-database for further usage.

\subsubsection{\textbf{Match the recently encoded photo with the encoded unrecognized photos that are saved in cross-directory of the system}}
While identifying a missing person, cross-matching is performed in the system. The system tries to find a match of the recently uploaded missing person's encoded photo with the collection of the unrecognized missing persons' encoded photos in the system. For the missing person's family side, the matching is performed against all the unrecognized encoded photos, uploaded by finders at different times. For the finder side, the matching will be performed against all the missing persons' encoded photos in the database which were uploaded by the missing persons' families at different times.

\subsubsection{\textbf{Show the matching result to the uploader}}
If the system gets any match between the recently uploaded photo and a prerecorded photo in the system, it shows the information of the other side to the user. For example, after getting a success on matching, the system shows the information of the finders, if the user is a family-member of the missing person and vice versa. Otherwise, it confirms users to be notified whenever any match would be found in the future.

\subsubsection{\textbf{Notify the appropriate person}}
The system notifies the family with the details of a finder of the missing person whenever the finder records any entry to the system. Again, a finder gets notified with the details of the family a missing person whenever the family records an entry about the missing person to the system.

\subsubsection{\textbf{Notify the police station}}
Whenever any finding happens, the applicable police station gets informed about the finding result so that the necessary steps from the police side can be performed immediately. 

\section{Results}\label{sec:result}
To test the functionalities of our system, \textit{postman} has been used as the front-end with some demo data conveying requests to the system that is working on the back-end.

The system's functionality can be investigated for the following cases:
\begin{description}
\item [\textbf{Case 1:}] \hspace{.2cm} A missing entry of a missing person is recorded first, then a finder records a finding entry of the person 

\item [\textbf{Case 2:}] \hspace{.2cm} Finding entry of a found person is recorded first, then family of the person records a missing entry of the person

\item [\textbf{Case 3:}] \hspace{.2cm} Already recorded missing entry is trying to be recorded multiple times 

\item [\textbf{Case 4:}] \hspace{.2cm} An intruder using \textit{false} info. while using the system
\end{description}
The results of our system in the cases mentioned above can be described as follows:\\
\subsection{Case 1: \textbf{A missing entry of a missing person is recorded first, then a finder records a finding entry of the person}}
\subsubsection{Recording a missing entry for missing person}
\begin{figure}[tbh]
   \frame{\includegraphics[width=3.5in, height=1.5in]{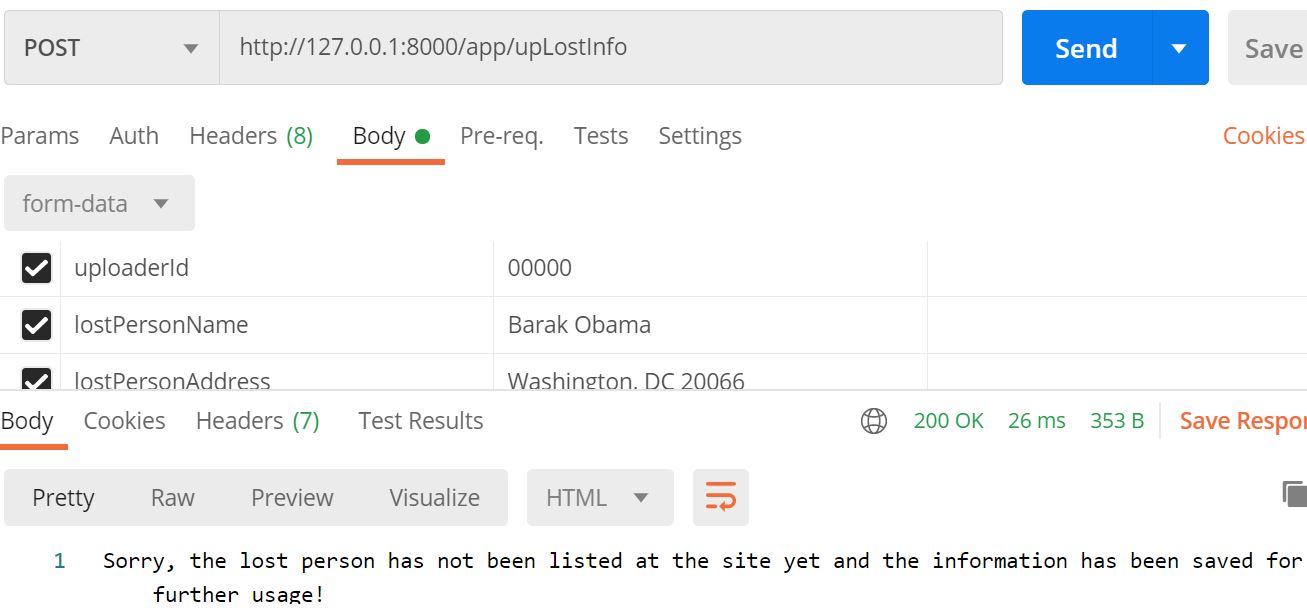}}
\caption{A missing entry}
\label{fig:fafimissing}
\end{figure}

In Fig~\ref{fig:fafimissing}, a missing person's photo along with valid information has been uploaded to the system first. After a successful verification, as there exists no previously listed finding record for the missing person in the cross directory of the system, the system will get no matching for the missing person. So, the information uploaded along with the photo of the missing person's information is stored in a specific directory for the future usage showing a message to the family member of the missing person as Fig.~\ref{fig:fafimissing}. 

\subsubsection{Recording a finding entry after finding the missing person}
\begin{figure}[tbh]
    \frame{\includegraphics[width=3.5in, height=1.5in]{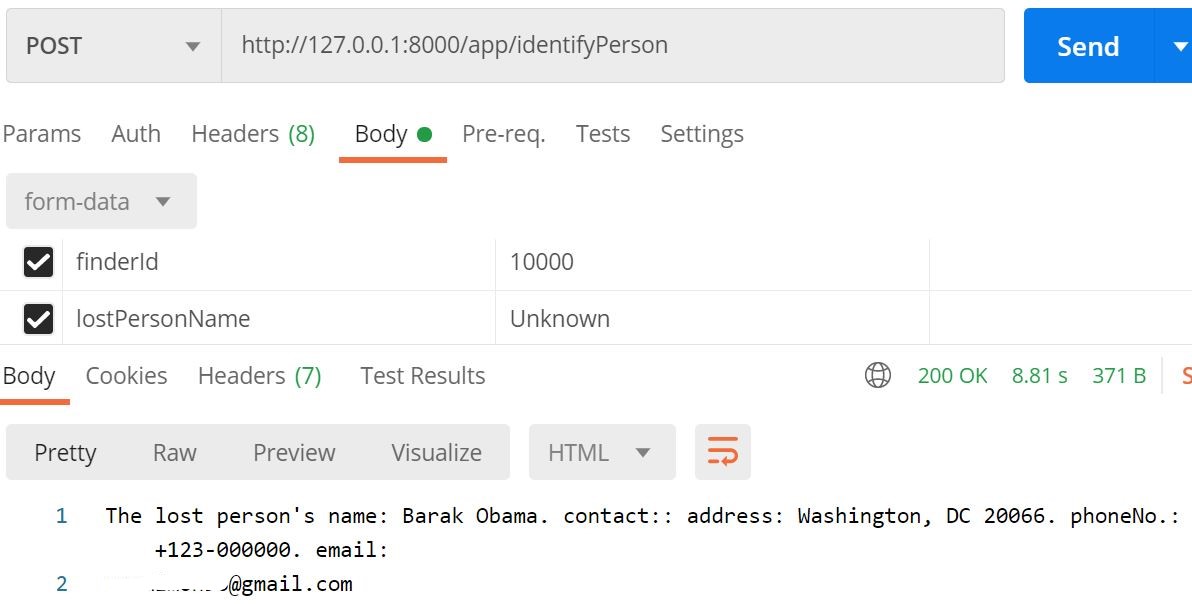}}
\caption{A finding entry}
\label{fig:fafifinding}
\end{figure}

Whenever a finder finds the missing person mentioned in the previous step and records a finding entry, after a proper verification, the system finds a successful match of the finder-uploaded photo with the photo uploaded by the family member in the previous step. Then the system shows the details of the family member of the missing person as Fig~\ref{fig:fafifinding} so that the finder can contact him easily.
\subsubsection{Notifying the concerns}
\begin{figure}[!tbh]
    \centering
    \subfloat[Email to missing person's family]{{\includegraphics[width=3.5in, height=1.5in]{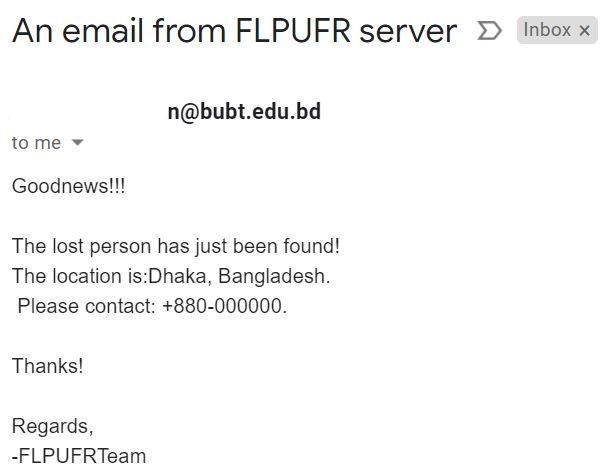} }}%
    \qquad
    \subfloat[Email to police station]{{\includegraphics[width=3.5in, height=1.5in]{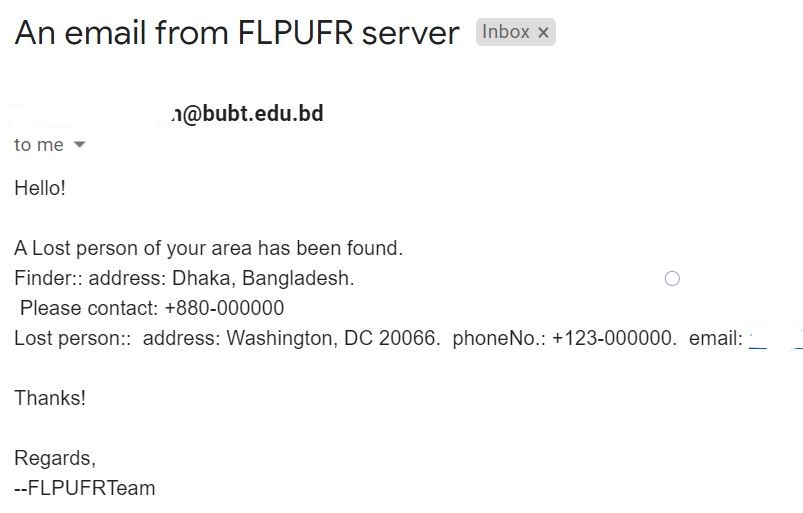} }}%
    \caption{Emails to the concerns}
    \label{fig:enamil1}
\end{figure}

In this case, the family of the missing person and the police station listed while recording the missing entry are notified via email about the matching success in the system. The email contains the details of the finder so that they can reach the finder avoiding any formality and hassle.
Sample emails -- sent to the family and the police station of the missing person's family side -- from the system are shown in Fig~\ref{fig:enamil1}.
\subsection{Case 2: \textbf{Finding entry of a found person is recorded first, then family of the person records a missing entry of the person}}
\subsubsection{Recording a finding entry after finding the missing person}
\begin{figure}[tbh]
   \frame{\includegraphics[width=3.5in, height=1.5in]{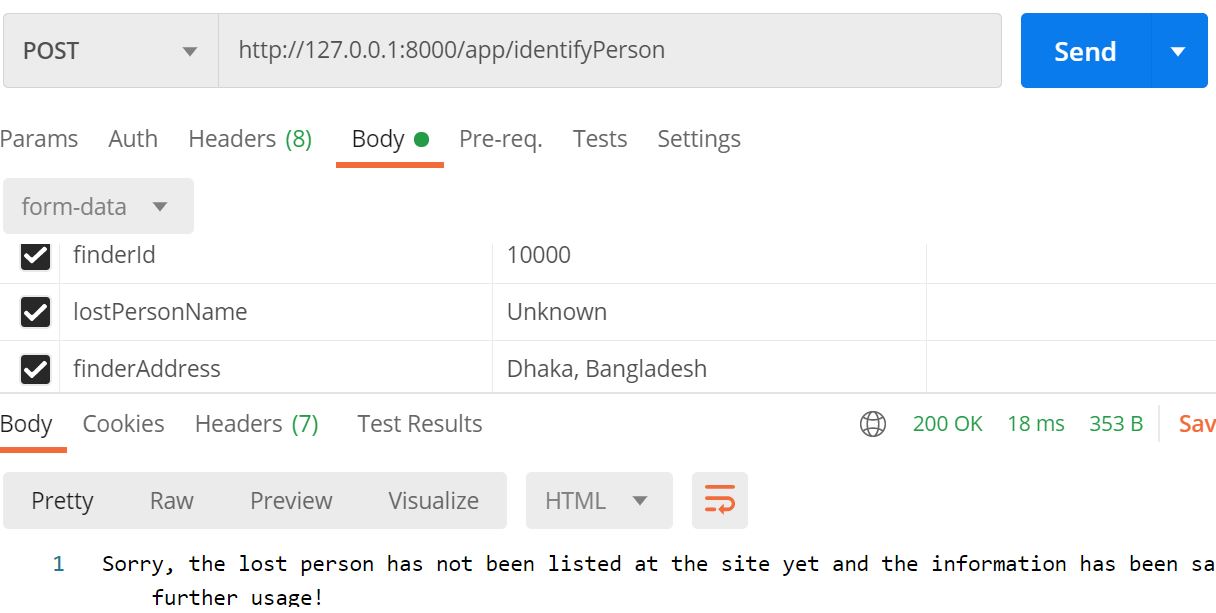}}
\caption{A finding entry}
\label{fig:fifafinding}
\end{figure}

In Fig~\ref{fig:fifafinding}, it can be seen that a finder, after finding a missing person, has recorded a finding entry with a verifiable photo of a missing person and the valid details of the finder. After a proper verification, the system fails to provide an instant solution to the finder regarding the missing person as the family member of the missing person has not recorded any entry yet to the cross directory of the system. That is why, the system saves the information for further usage and shows a message to the finder as Fig~\ref{fig:fifafinding}. 

\subsubsection{Recording a missing entry for the missing person}
\begin{figure}[tbh]
    \frame{\includegraphics[width=3.5in, height=1.5in]{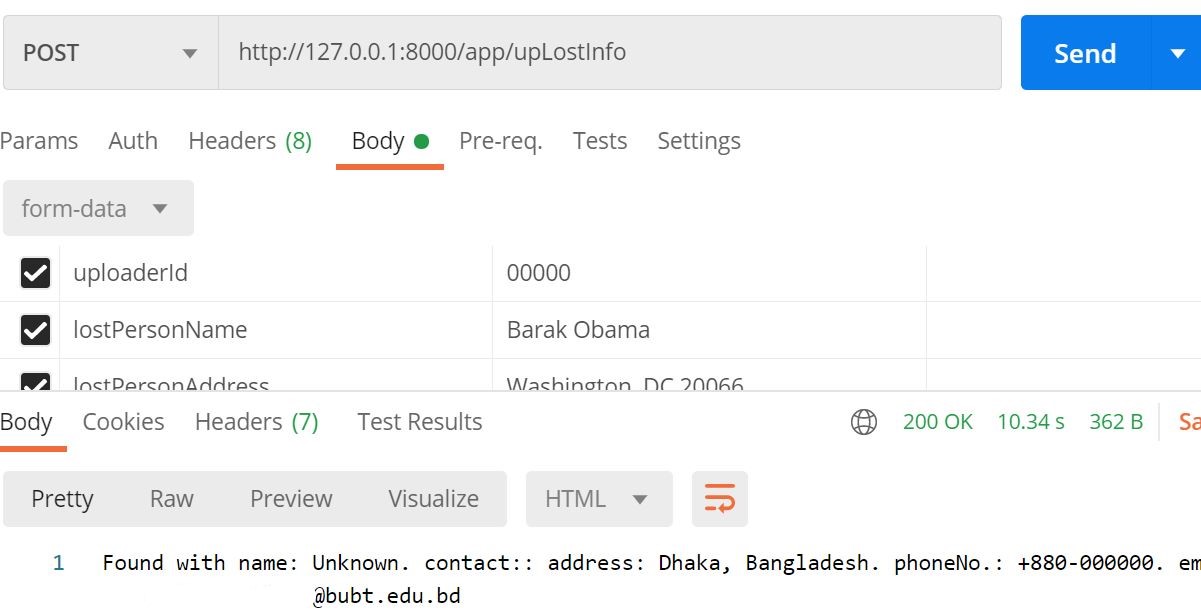}}
\caption{A missing entry}
\label{fig:fifamissing}
\end{figure}

Whenever the family of the person -- mentioned in the previous step -- records a missing entry to the system, after a successful verification, the system finds a match of the family-uploaded photo with the photo uploaded by the finder in the previous step and shows the details of the finder so that the family can contact him immediately. 
The scenario can be observed in Fig~\ref{fig:fifamissing}.
\subsubsection{Notifying the concerns}
\begin{figure}[!tbh]
    \centering
    \subfloat[Email to finder]{{\includegraphics[width=3.5in, height=1.5in]{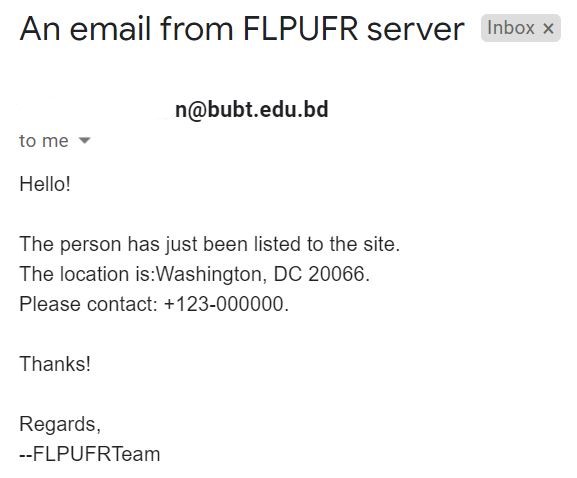} }}%
    \qquad
    \subfloat[Email to police station]{{\includegraphics[width=3.5in, height=1.5in]{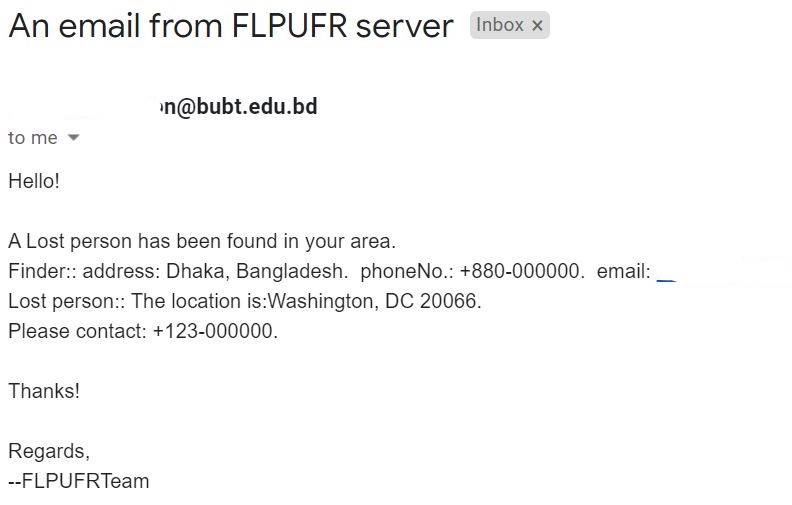} }}%
    \caption{Emails to the concerns}
    \label{fig:enamil2}
\end{figure}

In this case, the finder and the police station listed while recording the finding entry will be notified via email about the detail of the missing person so that they can reach the person's family. The email contains the details of the family-member of the missing person so that the finder can reach the the family-member avoiding any formality and hassle. Sample emails -- sent to the finder and the police station of the finder's side -- from the system are shown in Fig~\ref{fig:enamil2}.
\subsection{Case 3: \textbf{Already recorded missing entry is trying to be recorded multiple times}}
\begin{figure}[tbh]
    \frame{\includegraphics[width=3.5in, height=1.5in]{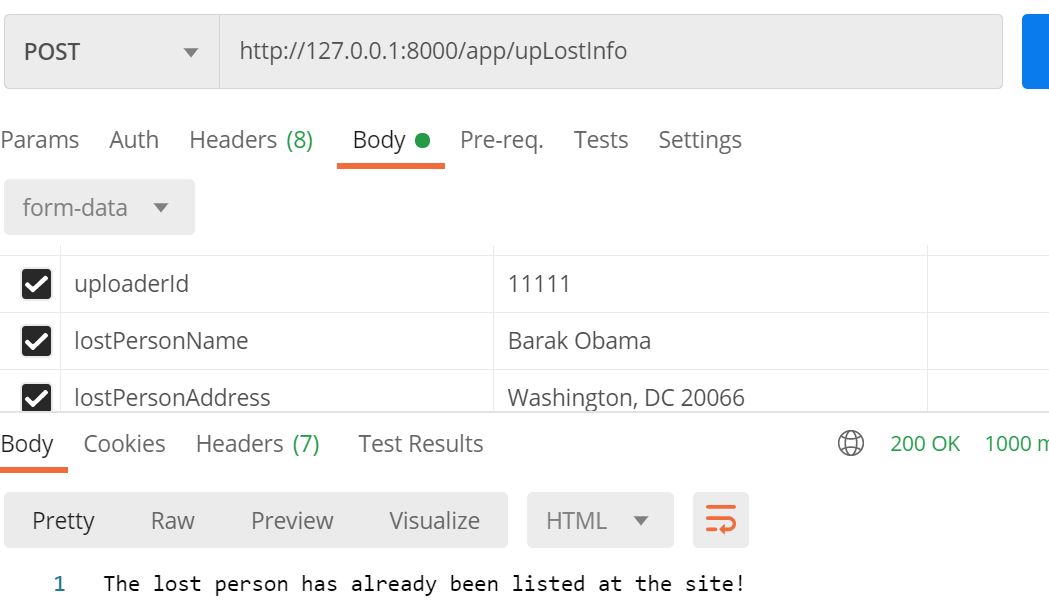}}
\caption{An attempt to record a prerecorded missing entry}
\label{fig:intruder}
\end{figure}

If an already recorded missing entry is trying to be recorded again and again, the databases inside the system might get populated by the redundant data which might cause a system failure. The task can be treated as an adversarial attempt to our system. Our system has the ability to avoid such adversarial tasks.\\
Fig~\ref{fig:intruder} shows that a person with a valid identity \textit{111111} has tried to record an entry that was recorded earlier. So after a proper verification, the system could detect and deny the task showing the user a short message. By this time, the system has also notified the family member and the police station -- listed while recording the first entry for the missing person -- about the attempt of recording the second entry for the missing person.

\subsection{Case 4: \textbf{An intruder using \textit{false} info. while using the system}}
\begin{figure}[tbh]
    \frame{\includegraphics[width=3.5in, height=1.5in]{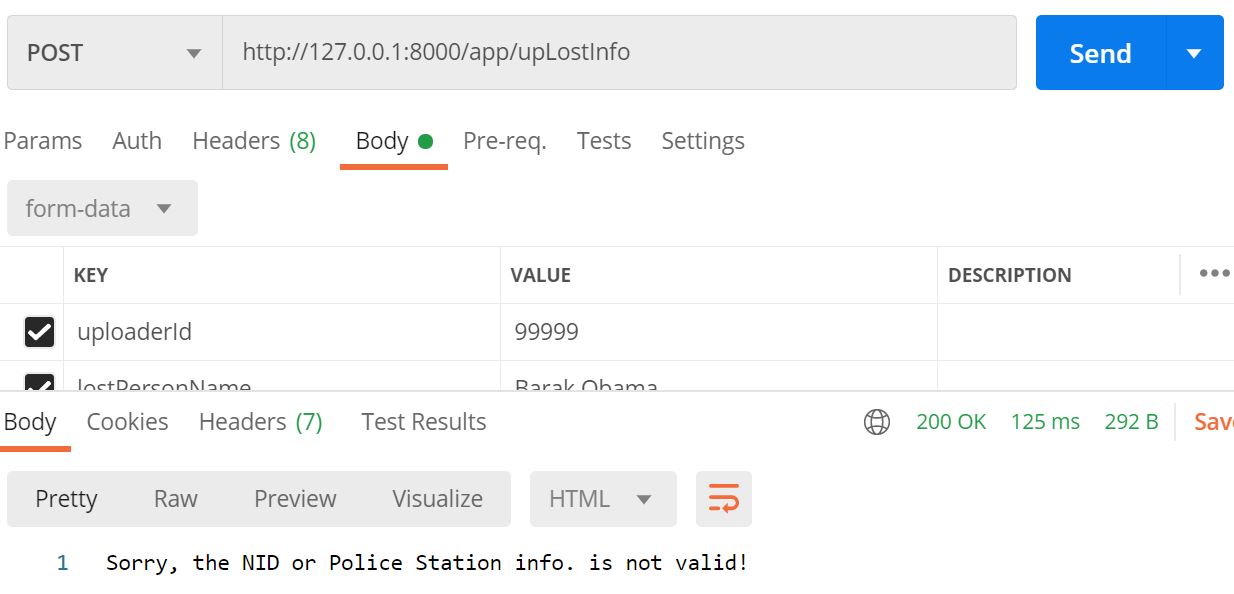}}
\caption{An intruder trying to use the system with false information}
\label{fig:intr}
\end{figure}

In case of finding a missing person, intruders can participate as unverified finders or family members of a missing person to take advantage of the situation. So, in the beginning, the users and the provided information have been verified thoroughly keeping the status of the procedure as \textit{AP} for security purposes.\\
A verification measure of our system can be observed in Fig~\ref{fig:intr}. Here an intruder is using a false NID \textit{99999} and other information. After the verification step, the system has found that the provided information is not valid and rejected the request of recording a missing entry to our system.
\section{Result Summary}\label{sec:resultsummary}
The main concern of the work was not on the facial recognition algorithm, instead, was on the security, feasibility, and efficiency. The facial recognition model of the work used a deep learning algorithm with an accuracy of 99.38\%\cite{eff} while performing the matching operation. The results of the work can be summarized and compared with other related works as in Table~\ref{tab1}.\\
\begin{table}[htbp]
\caption{Summary of the result compared to closely related works}\label{tab1}
\resizebox{\columnwidth}{!}{\begin{tabular}{|l|l|l|l|}
\hline
Benchmarks &  Our work & Reference work\cite{L1} & Reference work\cite{Ex}\\
\hline
Case 1 & Yes & Yes & Yes\\
\hline
Case 2 & Yes & No & Yes but Partially\\
\hline
Case 3 & Yes & Yes & No\\
\hline
Case 4 & Yes & No & No\\
\hline
Encryption & No & Yes & No\\
\hline
Cross directory checking & Yes & No & No\\
\hline
Technique used & Deep learning & FaceNet extension & SVM classifier\\
\hline
\end{tabular}}
\end{table}
Along with the security aspects, other characteristics such as equal importance to all of the stakeholders, feasibility, and efficiency ensure the completeness of the system. A unique feature -- maintaining and checking cross directory-- plays its role in sustaining efficiency while dealing with the data on a large scale. The area of consideration of our work is larger than any of the existing works. The services, provided by our system, are most effective and pragmatic.
\section{Future Works}\label{sec:fw}
In future, for security purposes of the system along with the missing persons, immutable digital identities~\cite{my} of the users will be used in our system. Moreover, The database will be switched from central to the distributed platforms~\cite{distributed} to ensure the security and ubiquity of the systems. Furthermore, hopefully, usage of real-time location would be integrated to the system for extending the existing functionalities of the system. The changes of a human's facial profile with the passage of time will also be considered in the next stages of our work.


\section{Conclusion}\label{sec:conclusion}
The incidents of missing people are not rare cases nowadays.
However, finding a missing person or helping others to find their missing member following an existing conventional way is somewhat complicated and vulnerable to intruders. People involved in the issue have to go through overbearing stress both physically and mentally. The system, proposed in this paper can provide a solution to the problem with the help of modern technology. Organization and the verification functionalities of the system play some crucial roles in developing a secure platform ensuring its efficiency in rigorous ambiance. From the results shown in the result section, it can be said that the system performs its functionalities with considerable perfection. The future extension of the work might grace the system with more stability, efficiency, and security resolving the dependency on the assumptions. Application of the system will provide an innovative and feasible solution to the problems related to missing persons and help them get off the complexities of the traditional solution of the problem.


\end{document}